\documentclass[conference]{IEEEtran}
\IEEEoverridecommandlockouts
\usepackage{cite}
\usepackage{algorithm}
\usepackage{booktabs}
\usepackage{amsmath,amssymb,amsfonts}
\usepackage{algpseudocode}
\floatname{algorithm}{\textbf{Algorithm}}
\usepackage{graphicx}
\usepackage{textcomp}
\usepackage{multirow}
\usepackage{xcolor}
\usepackage{amsmath}
\usepackage{amssymb}
\usepackage{array}
\usepackage{float}
\usepackage{subcaption}

\usepackage{caption}
\usepackage{bm}
\usepackage{wrapfig}
\usepackage{graphicx}
\usepackage{color,soul}
\def\KFA{\text{KnFu}}
\usepackage{adjustbox}
\usepackage{flushend}

\DeclareMathOperator*{\argmin}{argmin}
\author{\IEEEauthorblockN{S. Jamal Seyedmohammadi\IEEEauthorrefmark{1},
S. Kawa Atapour \IEEEauthorrefmark{2},
Jamshid Abouei\IEEEauthorrefmark{2},
Arash Mohammadi\IEEEauthorrefmark{1}
}

\IEEEauthorblockA{\IEEEauthorrefmark{1} Concordia Institute of Information Systems Engineering (CIISE), Concordia
University, Montreal, Canada}
\IEEEauthorblockA{\IEEEauthorrefmark{2} Dept. of Electrical Engineering, Yazd University, Yazd, Iran}

}

\def\BibTeX{{\rm B\kern-.05em{\sc i\kern-.025em b}\kern-.08em
    T\kern-.1667em\lower.7ex\hbox{E}\kern-.125emX}}

\begin{document}
\title{KnFu: Effective Knowledge Fusion\thanks{This work was partially supported by the Natural Sciences and Engineering Research Council (NSERC) of Canada through the
NSERC Discovery Grant RGPIN-2023-05654.}
}
\maketitle

\begin{abstract}
Federated Learning (FL) has emerged as a prominent alternative to the traditional centralized learning approach, attracting significant interest across a wide range of practical applications. Generally speaking, FL is a decentralized approach that allows for collaborative training of Machine Learning (ML) models across multiple local nodes, ensuring data privacy and security while leveraging diverse datasets. Conventional FL, however, is susceptible to  gradient inversion attacks, restrictively enforces a uniform architecture on local models, and suffers from model heterogeneity (model drift) due to non-IID local datasets. To mitigate some of these challenges, the new paradigm of Federated Knowledge Distillation (FKD) has emerged. FDK is developed based on the concept of Knowledge Distillation (KD), which involves extraction and transfer of a large and well-trained teacher model's knowledge to lightweight student models. FKD, however, still faces the model drift issue. Intuitively speaking, not all knowledge is universally beneficial due to the inherent diversity of data among local nodes. This calls for innovative mechanisms to evaluate the relevance and effectiveness of each client's knowledge for others, to prevent propagation of adverse knowledge. In this context, the paper proposes Effective Knowledge Fusion ($\KFA$) algorithm that evaluates knowledge of local models to only fuse semantic neighbors' effective knowledge for each client. The $\KFA$ is a personalized effective knowledge fusion scheme for each client, that analyzes  effectiveness of different local models' knowledge prior to the aggregation phase. Comprehensive experiments were performed on MNIST and CIFAR10 datasets illustrating effectiveness of the proposed $\KFA$ in comparison to its state-of-the-art counterparts. A key conclusion of the work is that  in scenarios with large and highly heterogeneous local datasets, local training could be preferable to knowledge fusion-based solutions. 

\end{abstract}

\section{\textbf{Introduction}}
\IEEEPARstart{F}{ederated} Learning (FL) has recently gained considerable attention, as an alternative to the centralized learning paradigm, in various domains including but not limited to computer vision, healthcare, and natural language processing. Generally speaking, FL resolves some practical challenges of centralized learning frameworks such as users' privacy issues and the communication cost of transmitting raw data from users/silos to the Fusion Centre (FC). Conventional FL methods aim to collaboratively train a global model by aggregating parameters of the clients' models without sharing their private data~\cite{FedAvg}. Such methods, however, pose the following new challenging problems: (i) Privacy concerns arising from gradient inversion attacks; (ii) Communication overhead of iterative transmission/reception of model parameters; (iii) Enforcing a uniform model architecture on clients, and; (iv) Model heterogeneity (model drift) resulted from non-IID local datasets~\cite{FedAvg challenges}.  

To mitigate some of the above mentioned challenges of conventional FL solutions,  the  new paradigm of Federated Knowledge Distillation (FKD)~\cite{FedMD} has been introduced that integrates the concept of Knowledge Distillation (KD) with FL. KD involves extracting the knowledge of a large and well-trained teacher model and transferring it to a lightweight student model by mimicking the teacher's predictions on a transfer set. In FKD, clients share only their local knowledge, i.e., predictions on the transfer set, with the server rather than their local model parameters. This leads to a more privacy-preserving framework, reduced communication overhead, and allowing heterogeneous model architectures among clients. While FKD presents effective advantages to resolve conventional FL's problems, it poses some new difficulties, including: (i) Requiring a transfer set to extract local knowledge of clients, and; (ii) Imposing  computation overhead on local devices. Additionally, the model drift issue still remains as an open challenge. 
Since clients hold non-IID local datasets, the models trained locally  would be heterogeneous, resulting in non-IID local knowledge among clients. Therefore, aggregating the local knowledge of a specific client with that of other clients may lead to adverse impacts on the client's local model resulting in significant performance degradation. Consequently, there has been a surge of recent interest devising innovative solutions to alleviate these issues~\cite{Adaptive KD, Generic Knowledge Distillation, Selective knowledge Sharing, Knowledge Selection, Clustered knowledge Transfer, Personalized Group Knowledge Transfer, MetaFed}. This field, however, is still in its infancy. The paper aims to further advance the research in this domain.   

\vspace{.025in}
\noindent
\textbf{Related Works:} In~\cite{Adaptive KD} an adaptive KD approach  is proposed, inspired by multitask learning methods, to adaptively adjust the weight of different distillation paths of an ensemble of teachers. Such an approach prevents negative impacts of some paths on the generalization performance of the student models. Reference~\cite{Generic Knowledge Distillation} studied whether all or partial knowledge of a model is effective. A generic knowledge selection method is presented to select and distill only certain knowledge by either fixing the knowledge selection threshold or changing it progressively during the training process as the teacher's confidence is enhanced. A selective knowledge-sharing mechanism is proposed in~\cite{Selective knowledge Sharing} to address the misleading and ambiguous knowledge fusion challenge resulting from non-IID local datasets and absence of a well-trained teacher model. The client-side selector chooses accurate predictions that match the ground-truth labels. Meanwhile, the server-side selector identifies the precise prediction by their entropy values. Precise knowledge has low entropy, while ambiguous predictions have high entropy and uncertainty.

Reference~\cite{Knowledge Selection} analyzed the effect of local predicted logits on the convergence rate. To improve the convergence rate, a knowledge selection method is proposed to schedule the predicted logits for efficient knowledge aggregation. In addition, a threshold-based approach is presented to optimize the local model updating options with/without knowledge distillation for each edge device to reduce the performance degradation of local models resulted from ambiguous knowledge. The COMET approach is proposed in~\cite{Clustered knowledge Transfer} introducing the clustered knowledge distillation concept, i.e., forming localized clusters from clients with similar data distribution. Each client then uses the aggregated knowledge of its cluster rather than following the average logits of all clients. Such an approach prevents  performance degradation by learning from clients with considerably different data distributions. In the local updating phase of each client, the loss function comprises a cross-entropy function along with a regularization term, which is an $l_2$-norm between the local and average predictions of the corresponding cluster. 
 
 KT-pFL~\cite{Personalized Group Knowledge Transfer} proposed a personalized group knowledge distillation algorithm, updating the personalized soft prediction of each client through a linear combination of all local predictions by a knowledge coefficient matrix. This matrix adaptively adjusts the collaboration among clients with similar data distribution and is parameterized to be trained simultaneously with the models. MetaFed \cite{MetaFed} presents a trustworthy personalized FL that achieves a personalized model for each federation without a central server using cyclic knowledge distillation. Its training process is split into two parts: common knowledge accumulation and personalization. In the first part, it leverages the validation accuracy on the current
federation’s validation data to decide whether to completely keep the previous federation’s knowledge and fine-tune it or just use it to update the current federation's through KD. In the personalization part, if the common model does not have enough performance on the validation data of the current federation, it refers little to it, while the weight of the KD regularization term is adapted if the common model’s performance is acceptable on the current validation data. 

\vspace{.025in}
\noindent
\textbf{Contributions:} The above mentioned works mainly focused to effectively distill knowledge of the teacher(s) into student model(s). In other words, the non-IID nature of local datasets has not yet been effectively addressed. Additionally, effectiveness of local knowledge of clients has not yet been investigated. The paper addresses these gaps, by development of a more efficacious knowledge fusion technique, aiming to present more thorough evaluation on the effectiveness of the local knowledge of clients. Our main contributions can be summarized as follows:
\begin{itemize}
\item  Proposal of the $\KFA$ algorithm that strategically evaluates and fuses only relevant and beneficial knowledge among clients. This personalized approach ensures that knowledge fusion is tailored to the semantic neighbors of each client, mitigating the risk of model drift caused by non-IID local datasets.
\item Introduction of a novel mechanism within the $\KFA$ algorithm to assess the relevance and impact of shared knowledge across clients, ensuring that only effective knowledge contributes to the FL process, thereby preventing the dilution of model performance with non-contributory information.
%
\end{itemize}
Comprehensive experiments were performed on MNIST, and CIFAR10 datasets to show the effectiveness of the proposed algorithm in comparison with baseline methods in terms of different metrics. The rest of the paper is organized as follows: Section~\ref{sec:PS} formulates the problem and provides required background material to follow developments of the papers.  The  $\KFA$ algorithm is proposed in Section~\ref{sec:KFA}, while Section~\ref{sec:EX} presents simulation results and analysis. Finally, Section~\ref{sec:CONC} concludes the paper.

\section{\textbf{Preliminaries and Problem Statement}} \label{sec:PS}
\begin{figure}[t!]
\centering
\begin{subfigure}{0.22\textwidth}
\includegraphics[width=\textwidth]{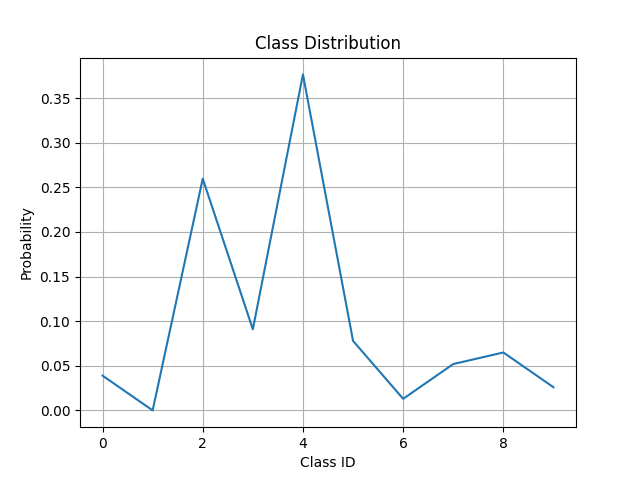}
\caption{Class Distribution}
\end{subfigure}
\begin{subfigure}{0.22\textwidth}
\includegraphics[width=\textwidth]{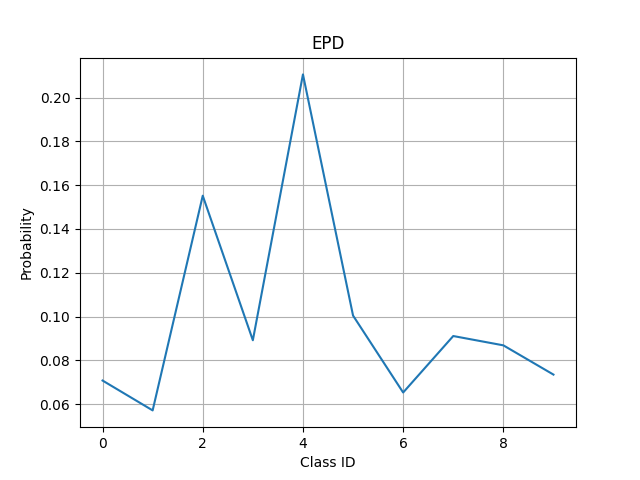}
\caption{EPD}
\end{subfigure}
\caption{\footnotesize The EPD as an estimation of the class distribution. \label{Fig:EPD}}
\end{figure}
\begin{figure*}[t!]
\centering
\begin{subfigure}{0.32\textwidth}
\includegraphics[width=\textwidth]{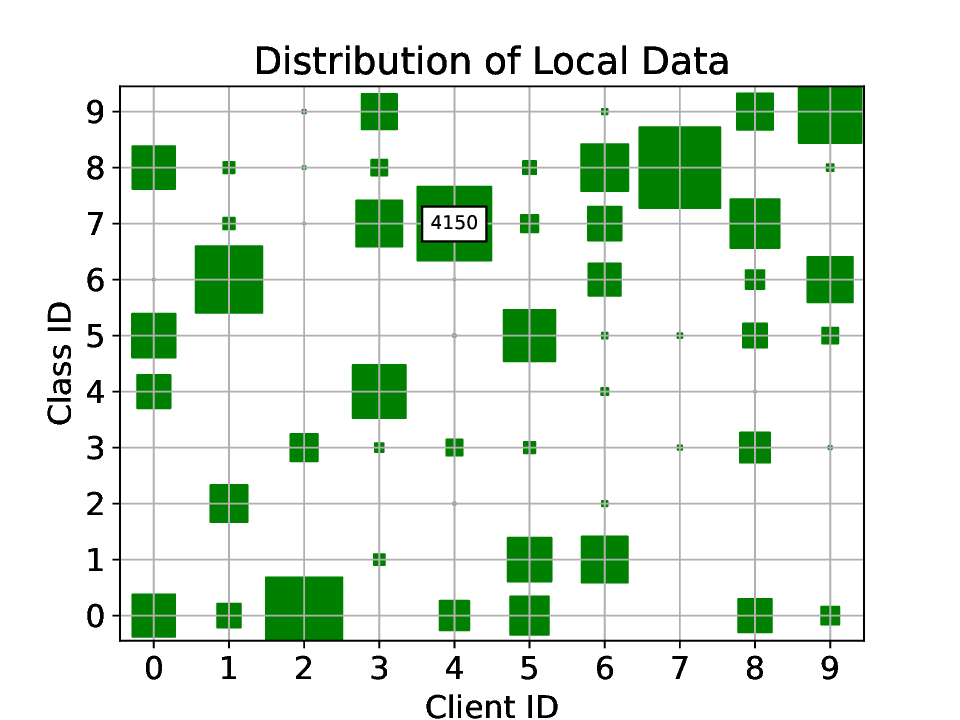}
\caption{$\alpha=0.1$}
\end{subfigure}
\hfill
\begin{subfigure}{0.32\textwidth}
\includegraphics[width=\textwidth]{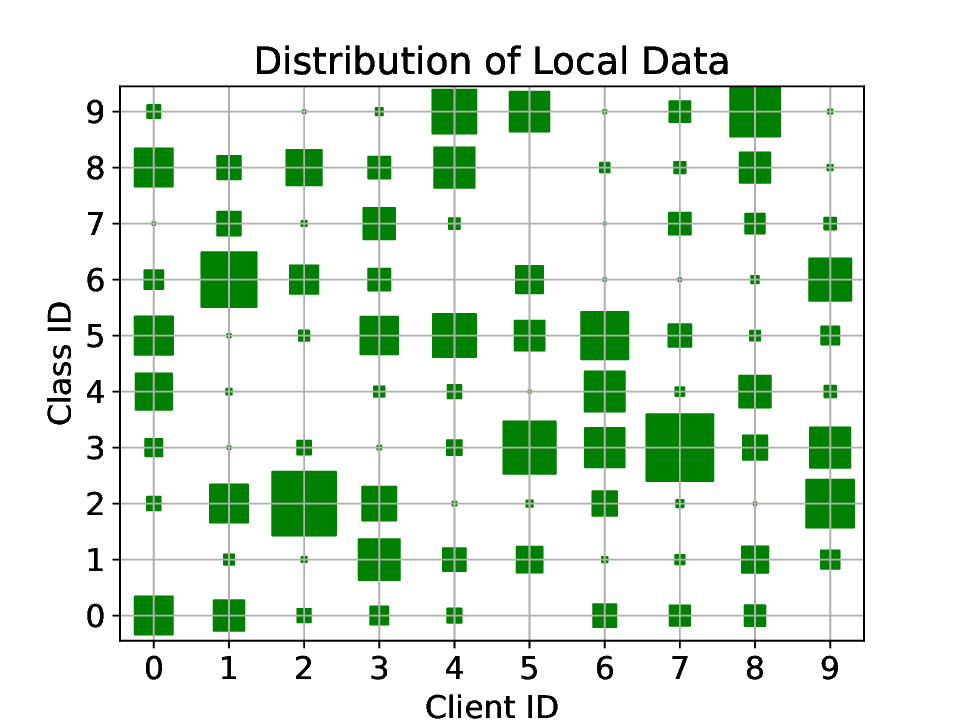}
\caption{$\alpha=0.5$}
\end{subfigure}
\hfill
\begin{subfigure}{0.32\textwidth}
\includegraphics[width=\textwidth]{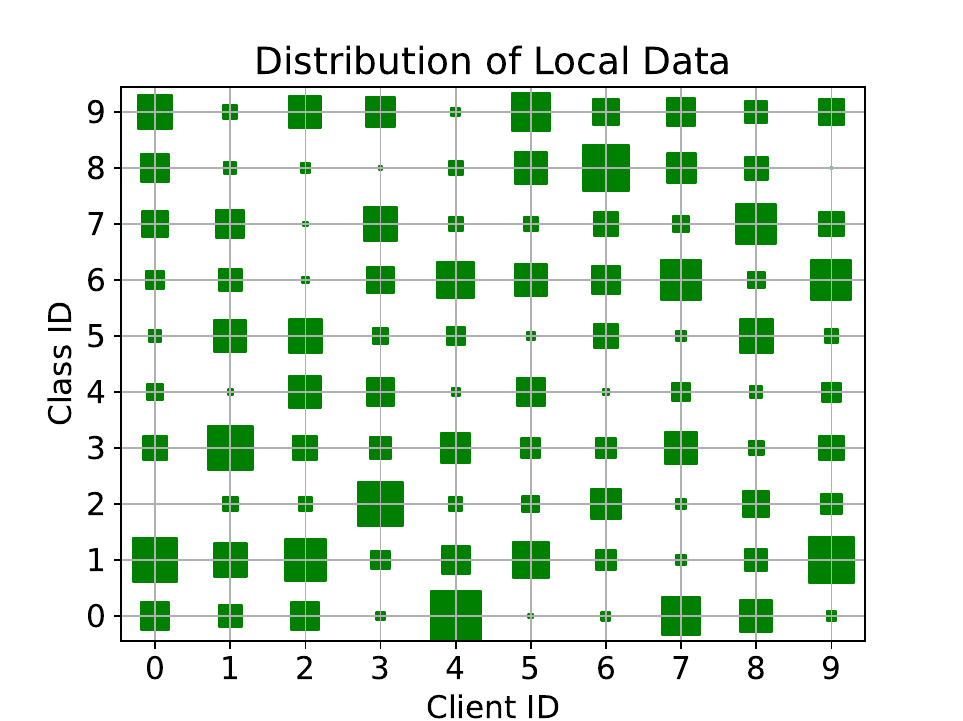}
\caption{$\alpha=1$}
\end{subfigure}
\caption{\footnotesize Illustration of data heterogeneity among $10$ clients on the CIFAR-10 dataset.} \label{fig_data_het}
\end{figure*}

In this paper, we aim to perform a supervised $C$-class classification task.
Let's consider a set of $N$ clients, denoted by $\mathbb{U}=\{u_1, ...,u_N\}$, in the FL system coordinated by a fusion center. Local datasets of clients, represented by $\mathbb{D}_n = \bigcup_{i=1}^{K_n} \{(x_n^i, y_n^i)\}, n \in \{1, ..., N\}$, are heterogeneous, where $(x_n^i, y_n^i)$ denotes $i^{th}$ data sample including the input and its ground-truth output, and $K_n$ indicates the size of the local dataset $\mathbb{D}_n$.
Each client $u_n$ aims to train a local Convolutional Neural Network (CNN) model, denoted by $f(\cdot; \boldsymbol\Omega_n)$, parameterized by $\boldsymbol\Omega_n$. Let $f(x; \boldsymbol\Omega) = [f_1(x; \boldsymbol\Omega), ... ,f_C(x; \boldsymbol\Omega)]$ denote the output of the last layer (softmax) of the CNN model, where $\sum_{j=1}^{C}f_j(x; \boldsymbol\Omega) = 1$, and $f_j(x; \boldsymbol\Omega)$ indicates the probability of assigning data sample $\boldsymbol{x}$ to the $j^{th}$ class. 

Term $\boldsymbol{p}(\boldsymbol{\Omega}) = [p_1(\boldsymbol{\Omega}), ..., p_{C}(\boldsymbol{\Omega})]$ is defined as the Estimated Probability Distribution (EPD) of assigning new data samples to different classes. EPD is calculated over a shared dataset among clients, the transfer set, with $K_t$ data samples, denoted by $\mathcal{D}^t = \bigcup_{i=1}^{K_t} \{(x_t^i, y_t^i)\}$. To investigate the local model's bias to different classes, EPD can be computed as the expectation of probability distribution of the local models on the data samples of the transfer set, as follows
\begin{equation}
\label{average_probability_distribution}
\boldsymbol p(\boldsymbol{\Omega}) = \sum_{x\in \mathcal{D}^t}\frac{f(x;\boldsymbol{\Omega})}{|\mathcal D^t|}.
\end{equation}

KD involves extracting and transferring knowledge from a well-trained teacher model into a student model, emulating the outputs of the teacher model using a transfer set. Specifically, a Kullback-Leibler (KL) divergence function~\cite{KD} is utilized to minimize the discrepancy between the soft labels of the teacher model and the student model, as follows 
\begin{equation}
\begin{aligned}\label{KD}
\boldsymbol\Omega^{*}_s &= \argmin_{\boldsymbol\Omega_s} \mathbb{E}_{(x,y)\sim \mathcal{D}^t} \bigg\{\mathcal{L}_{CE}(f( x; \boldsymbol\Omega_s), y)\\
&+ \lambda^2 \mathcal{L}_{KL} (f(x; \boldsymbol\Omega_s), f( x; \boldsymbol\Omega_t)) \bigg\},
\end{aligned}
\end{equation}
where $f(x; \boldsymbol\Omega_s)$ and $f( x; \boldsymbol\Omega_t)$ denote the predictions of the student and teacher models for input $x$, respectively. In addition, $\mathcal{L}_{CE}$ and $\mathcal{L}_{KL}$ are the cross-entropy and KL loss functions, and $\lambda$ is the so-called temperature hyper-parameter used to soften generated logits.
In~\cite{FedMD}, clients first update their local models using their respective local datasets, then, each client performs predictions on the transfer set to extract a set of soft labels, known as local knowledge. These local soft labels are averaged in the fusion center to fuse the local knowledge of the clients. Finally, the average soft labels, known as collaborative knowledge, are utilized in Eq~\eqref{KD} to distill it into the local models. In~\cite{FedMD} and similar papers, the local knowledge of all clients is aggregated and distilled to each local model. The local models are, however, heterogeneous due to the non-IID nature of local datasets. Consequently, their local knowledge is also heterogeneous, i.e.,  local knowledge of a specific client may not be effective for all other clients in the FL system. 

To address the above mentioned issue, in this paper, we aim to answer the followings questions: \textit {Q1: How can we assess the effectiveness of a specific client's local knowledge for other clients? Q2: How can we transfer only the effective knowledge and ignore the adverse knowledge of local models?} To answer the first question (Q1), we assess the effectiveness of knowledge-sharing and local training options based on two important factors: $(i)$ Data heterogeneity level, and; $(ii)$ Local dataset size. To answer the second question, we propose the innovative Effective Knowledge Fusion ($\KFA$) algorithm that evaluates knowledge of local models and fuse semantic neighbors' (i.e., clients with similar data distributions) effective knowledge for each client.  

\section{\textbf{The Proposed $\KFA$ Algorithm}} \label{sec:KFA}
In this section, we present the proposed $\KFA$ algorithm that effectively combines useful knowledge of various clients within an FL framework, ultimately leading to effective personalized local models. Given the inherent diversity of data among clients, it is crucial to acknowledge that not all knowledge is universally beneficial. Hence, there is a necessity for a mechanism to evaluate the relevance and effectiveness of each client's knowledge for others, therefore, preventing the propagation of adverse knowledge. Moreover, crafting an efficient methodology to distill useful knowledge into specific local models is challenging. The proposed $\KFA$ algorithm, consisting of four primary steps, operates over $R$ rounds or until convergence is achieved.

\vspace{.025in}
\noindent
\textbf{\textit{Step 1: Local Training:}} Initially, individual models undergo training on their local datasets for a set number of local epochs, denoted by $E$, as follows
\begin{equation}\label{pFL}
\boldsymbol\Omega_n^*= \argmin_{\boldsymbol\Omega_n}\mathbb{E}_{(x,y)\sim\mathbb D_n}\{\mathcal{L}_{CE} (f( x;\boldsymbol\Omega_n), y)\}.
\end{equation}

\noindent
\textbf{\textit{Step 2: Local Knowledge Extraction:}} Following the update of local models, their knowledge is extracted via the transfer set. This extraction involves obtaining soft labels from each local model for the data samples within the transfer set, i.e.,
\begin{equation}\label{KE}
\boldsymbol F_n = f( x;\boldsymbol\Omega_n), \forall x \in \mathcal D^t,
\end{equation}
where $\boldsymbol F_n$ constitutes a matrix where each row corresponds to a data sample within the transfer set. The columns of this matrix show the probability distribution for assigning a particular data sample to various classes. 

\noindent
\textbf{\textit{Step 3: Effective Knowledge Fusion:}} The knowledge extracted from individual clients is transferred to the fusion center for aggregation, leading to the creation of personalized fused knowledge for each client. Initially, we calculate the Estimated Probability Distributions (EPDs) for clients, which indicate the bias of their local models towards various classes. As illustrated in Fig.~\ref{Fig:EPD}, EPD serves as an estimation of the class distribution within each client's dataset,  and is 
\begin{eqnarray}\label{EPD}
\boldsymbol p(\boldsymbol{\Omega}_n) = \frac{1}{K_t}\sum_{i = 1}^{K_t}\boldsymbol F_n^{i,j}, \forall n \in \{1, \ldots, N\},
\end{eqnarray}
where $\boldsymbol F_n^{i,j}$ indicates the probability of assigning the $i^{th}$ data sample in the transfer set to the $j^{th}$ class using the local model of user $u_n$.
Next, we adjust the weighting of each client's local knowledge for a specific client according to how similar their EPDs are to that of the client. We measure the distance between two distributions using KL divergence, calculated as 
\begin{equation}\label{KL}
d_{n,m} = KL \Big(\boldsymbol p(\boldsymbol{\Omega}_n), \boldsymbol p(\boldsymbol{\Omega}_m)\Big), \forall n,m \in \{1, ..., N\}.
\end{equation}
To determine the importance of the knowledge from client $u_m$ for the client $u_n$, we evaluate the similarity between their EPDs by taking the inverse of the squared distance calculated in Eq. (\ref{KL}), as follows
\begin{equation}\label{weight}
w_{n,m} = \frac{1}{\Big(d_{n,m}\Big)^2}, \forall n,m \in \{1, \ldots, N\}.
\end{equation}
Notably, the weight of the local knowledge is adjusted by a positive constant $\beta$, as
\begin{equation}\label{beta}
w_{n,n} = \beta \times \max \{w_{n,m}\}_{m=1}^N.
\end{equation}
Finally, the personalized aggregated knowledge for client $u_n$ is calculated as
\begin{equation}\label{pFL}
\boldsymbol F_n^{agg} = \sum_{m=1}^{N}\frac{w_{n,m}}{\sum_m w_{n,m}}\times\boldsymbol F_m, \forall n \in \{1, ..., N\}.
\end{equation}

\vspace{.025in}
\noindent
\textbf{\textit{Step 4: Local Model Fine-tuning:}} In the final step, the personalized fused knowledge is distributed to clients, allowing them to integrate effective knowledge from other clients into their local models. Clients refine their local models by incorporating the aggregated knowledge alongside their local datasets during fine-tuning via the transfer set,  as follows
\begin{equation}
\begin{aligned}
\boldsymbol\Omega^{*}_n &= \argmin_{\boldsymbol\Omega_n} \mathbb{E}_{(x,y)\sim \mathcal{D}^t} \bigg\{\mathcal{L}_{CE}(f( x; \boldsymbol\Omega_n), y)\\
&~~~~~~~~~~~~~~~~+ \lambda^2 \mathcal{L}_{KL} (f(x; \boldsymbol\Omega_n), \boldsymbol F_n^{agg}) \bigg\},
\end{aligned}
\end{equation}
where $\mathcal{L}_{KL}$ denotes the KL loss function and $\lambda$ adjusts the balance between the two terms of the loss function. The first term focuses only on the local dataset, while the second term concentrates on the aggregated knowledge of clients. This completes description of the proposed $\KFA$, next we present our simulation results and analysis.

\begin{algorithm}[t!]
\renewcommand{\thealgorithm}{1}
\caption{Pseudocode of the proposed $\KFA$ algorithm}
\label{alg1}
\begin{algorithmic}
\State \textbf{Input:} Local datasets $\{\mathbb D_n\}_{n=1}^N$ and transfer set $\mathcal D^t$.

\For{$r = 1, \dots, R$}
\State \textit{\#\#  Local Training}

\For{$u_n \in \mathbb U$}
\State $\boldsymbol\Omega_n^*= \argmin_{\boldsymbol\Omega_n}\mathbb{E}_{(x,y)\sim\mathbb D_n}\{\mathcal{L}_{CE} (f( x;\boldsymbol\Omega_n), y)\}$
\EndFor

\State \textit{\#\#  Local Knowledge Extraction}

\For{$u_n \in \mathbb U$}
\State $\boldsymbol F_n = f( x;\boldsymbol\Omega_n), \forall x \in \mathcal D^t$
\EndFor

\State \textit{-------------------------------------------FUSION CENTER--------}
\State \textit{\#\# Effective Knowledge Fusion}
\For{$u_n \in \mathbb U$}

\State \textit{\#\# estimation of class distribution (EPD)}
\State $\boldsymbol p(\boldsymbol{\Omega_n}) = \frac{1}{K_t}\sum_{i = 1}^{K_t}\boldsymbol F_n^{i,j}$

\State \textit{\#\# dissimilarities between EPDs}
\State $d_{n,m} = KL \Big(\boldsymbol p(\boldsymbol{\Omega}_n), \boldsymbol p(\boldsymbol{\Omega}_m)\Big)$

\State \textit{\#\# personalized knowledge fusion}
\State $\boldsymbol F_n^{agg} = \sum_{m=1}^{N}\frac{w_{n,m}}{\sum_m w_{n,m}}\times\boldsymbol F_m$
\EndFor

\State \textit{------------------------------------------------------------------------}

\State \textit{\#\#  Local Model Fine-tuning}
\For{$u_n \in \mathbb U$}

\State \textit{Eq. (10)}
\EndFor
\EndFor
\State \textbf{Output:} Personalized local models $\{\boldsymbol\Omega_n\}_{n=1}^N$ 
\end{algorithmic}
\end{algorithm}
\begin{figure*}[t!]
	\centering
	\begin{subfigure}{0.32\textwidth}
		\includegraphics[width=\textwidth]{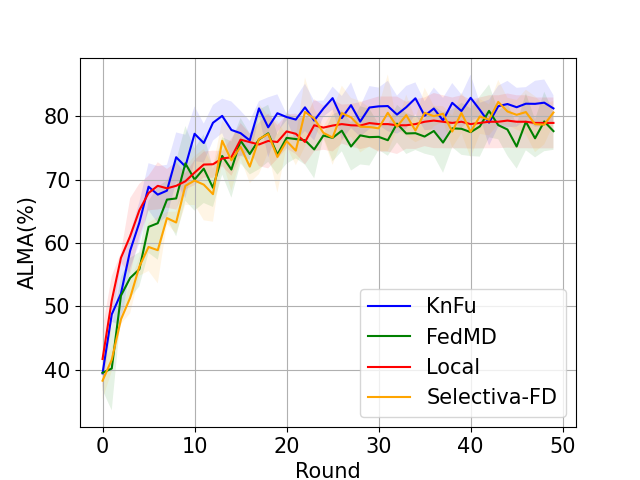}
		\caption{$|\mathbb D_n| = 50$}
	\end{subfigure}
	\begin{subfigure}{0.32\textwidth}
		\includegraphics[width=\textwidth]{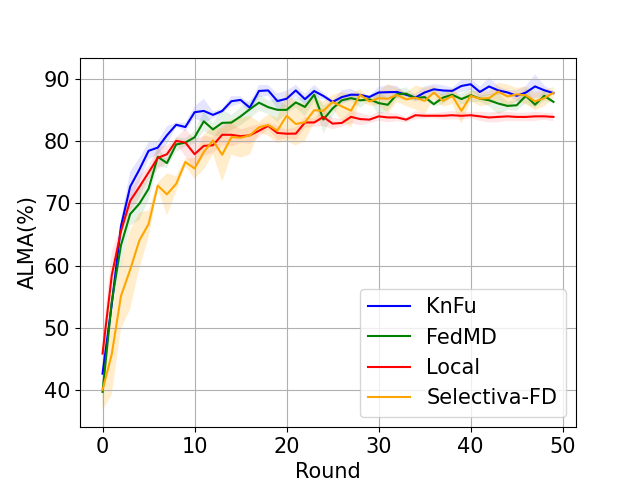}
		\caption{$|\mathbb D_n| = 100$}
	\end{subfigure}
	\begin{subfigure}{0.32\textwidth}
		\includegraphics[width=\textwidth]{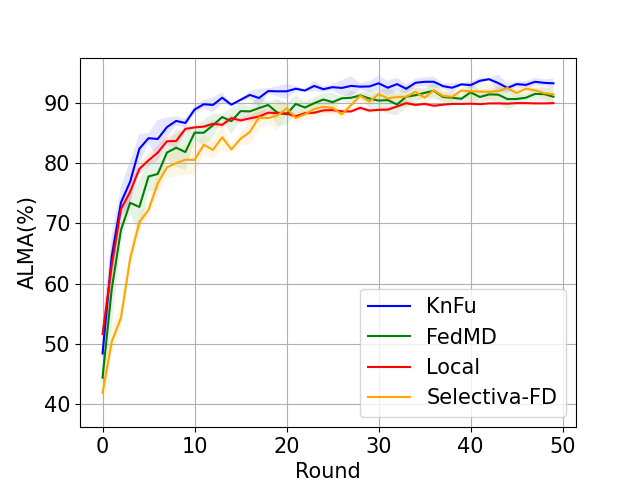}
		\caption{$|\mathbb D_n| = 200$}
	\end{subfigure}
	\caption{\footnotesize Learning curves of the ALMA metric corresponding to different methods, for a fixed heterogeneity level, $\alpha= 0.5$, and different local data sizes, $|\mathbb D_n| = \{50, 100, 200\}$. } \label{local data size}
\end{figure*}
\begin{figure*}[t!]
\centering
\begin{subfigure}{0.38\textwidth}
\includegraphics[width=\textwidth]{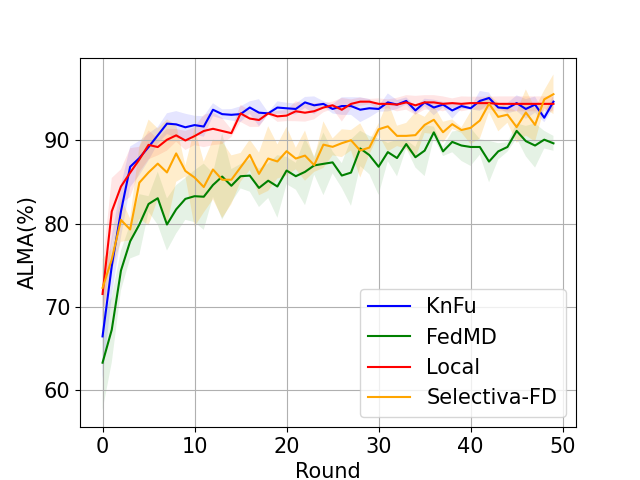}
\caption{$\alpha = 0.1$}
\end{subfigure}
\hspace{2cm}
\begin{subfigure}{0.38\textwidth}
\includegraphics[width=\textwidth]{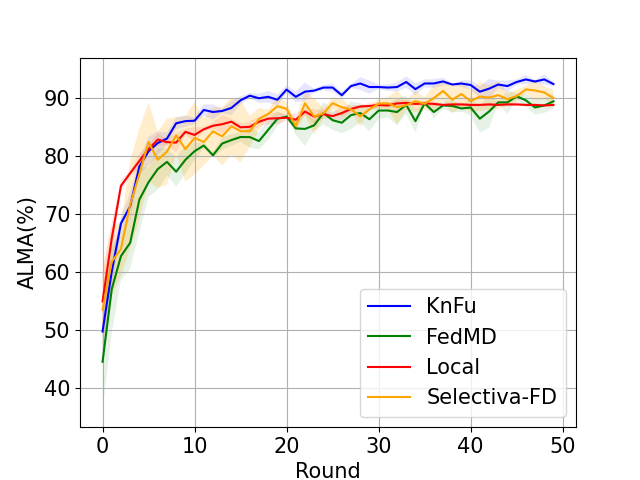}
\caption{$\alpha = 0.25$}
\end{subfigure}
\begin{subfigure}{0.38\textwidth}
\includegraphics[width=\textwidth]{mnist_alpha50_D100.png}
\caption{$\alpha = 0.5$}
\end{subfigure}
\hspace{2cm}
\begin{subfigure}{0.38\textwidth}
\includegraphics[width=\textwidth]{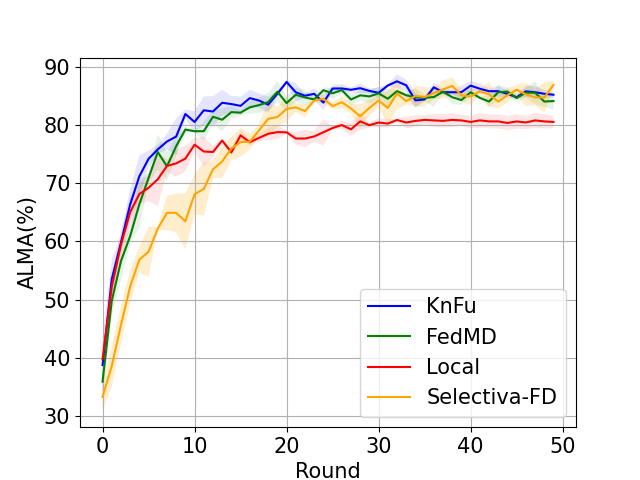}
\caption{$\alpha = 1$}
\end{subfigure}
\caption{\footnotesize Learning curves of the ALMA metric associated with different methods, for a fixed local data size, $|\mathbb D_n| = 100$, and various heterogeneity levels, $\alpha= \{0.1, 0.25, 0.5, 1$\}.} \label{heterogeneity level}
\end{figure*}
\begin{center}
\begin{table*}[t]
\caption{ALMA (\%) given different data settings on MNIST dataset. \label{MNIST_table} }
 \begin{adjustbox}{width=\textwidth}
\begin{tabular}[t]{ m{1cm} m{6.2cm} m{6.2cm}  m{6.2cm} }
\hline
Het. Level&                                  \begin{tabular}{c} $|\mathbb D_n| = 50$  \\ \hline \hspace{0.1cm} $\KFA$ \hspace{0.5cm} FedMD \hspace{0.5cm} Local \hspace{0.5cm} Selective-FD \end{tabular}&        \begin{tabular}{c} $|\mathbb D_n| = 100$ \\ \hline  \hspace{0.1cm} $\KFA$ \hspace{0.5cm} FedMD \hspace{0.5cm} Local \hspace{0.5cm} Selective-FD \end{tabular}&      \begin{tabular}{c} $|\mathbb D_n| = 200$  \\ \hline \hspace{0.1cm} $\KFA$ \hspace{0.5cm} FedMD \hspace{0.5cm} Local \hspace{0.5cm} Selective-FD \end{tabular}\\
\hline
$\alpha=0.1$\newline$\alpha=.25$\newline$\alpha=0.5$\newline$\alpha=1$&       
\underline{$93.5\pm0.6$} \hspace{0.2cm} $88.7\pm0.6$\hspace{0.2cm} $92.4\pm2.7$ \hspace{0.2cm} $92.2\pm1.3$  \newline  \underline{$90.4\pm1.3$} \hspace{0.2cm} $86.0\pm1.1$\hspace{0.2cm} $88.9\pm2.1$ \hspace{0.2cm} $88.5\pm1.7$ \newline 
\underline{$81.5\pm2.7$} \hspace{0.2cm} $78.1\pm3.6$\hspace{0.2cm} $79.0\pm4.0$ \hspace{0.2cm} $79.8\pm2.6$ \newline 
$78.5\pm3.9$ \hspace{0.2cm} $76.5\pm3.3$\hspace{0.2cm} $75.3\pm5.0$ \hspace{0.2cm} $\underline{78.9\pm4.1}$&                   $94.1\pm0.4$ \hspace{0.2cm} $89.3\pm0.9$\hspace{0.2cm} \underline{$94.4\pm0.8$} \hspace{0.2cm} $93.1\pm1.5$  \newline  \underline{$92.3\pm0.6$} \hspace{0.2cm} $88.6\pm1.2$\hspace{0.2cm} $88.7\pm0.2$ \hspace{0.2cm} $90.3\pm1.1$\newline 
\underline{$88.1\pm0.6$} \hspace{0.2cm} $86.4\pm0.2$\hspace{0.2cm} $83.9\pm0.6$ \hspace{0.2cm} $87.1\pm1.0$ \newline 
\underline{$85.6\pm0.1$} \hspace{0.2cm} $84.9\pm0.6$\hspace{0.2cm} $80.5\pm1.1$\hspace{0.2cm} $85.2\pm1.0$&        
$96.2\pm0.4$ \hspace{0.2cm} $92.2\pm0.8$\hspace{0.2cm} \underline{$96.3\pm0.6$}\hspace{0.2cm} $95.9\pm0.6$   \newline  \underline{$94.4\pm0.7$} \hspace{0.2cm} $92.0\pm1.3$\hspace{0.2cm} $92.6\pm1.1$ \hspace{0.2cm} $93.0\pm0.9$ \newline 
\underline{$93.3\pm0.5$} \hspace{0.2cm} $91.2\pm1.1$\hspace{0.2cm} $89.9\pm0.0$ \hspace{0.2cm} $91.9\pm0.9$ \newline 
$92.1\pm0.8$ \hspace{0.2cm} $92.0\pm0.4$\hspace{0.2cm} $87.8\pm0.8$\hspace{0.2cm} \underline{$92.2\pm0.8$}\\
\hline
\end{tabular}
 \end{adjustbox}
 
\end{table*}
\end{center}
\begin{center}
\begin{table*}[ht]\centering 
\caption{ALMA (\%) given different data settings on CIFAR-10 dataset. \label{CIFAR_table}}
\begin{tabular}[t]{ m{1cm} m{7cm} m{7cm} }
\hline
Het. Level&                                  \begin{tabular}{c} $|\mathbb D_n| = 500$  \\ \hline \hspace{0.1cm} $\KFA$ \hspace{0.8cm} FedMD \hspace{0.8cm} Local \hspace{0.8cm} Selective-FD \end{tabular}&        \begin{tabular}{c} $|\mathbb D_n| = 1000$  \\ \hline  \hspace{0.1cm}$\KFA$ \hspace{0.8cm} FedMD \hspace{0.8cm} Local \hspace{0.8cm} Selective-FD  \end{tabular}\\
\hline
$\alpha=0.1$\newline$\alpha=0.5$\newline$\alpha=1$&              $72.00\pm1.57$ \hspace{0.2cm} $50.50\pm2.51$\hspace{0.2cm} \underline{$77.80\pm1.78$}  \hspace{0.2cm} $73.34\pm2.47$ \newline  $48.41\pm2.30$ \hspace{0.2cm} $41.57\pm1.94$\hspace{0.2cm} \underline{$50.70\pm1.25$} \hspace{0.2cm} $48.07\pm2.13$ \newline \underline{$45.30\pm2.49$} \hspace{0.2cm} $42.20\pm2.73$\hspace{0.2cm} $43.68\pm1.83$\hspace{0.2cm} $44.38\pm2.63$&                   $73.80\pm1.23$ \hspace{0.2cm} $52.34\pm1.18$\hspace{0.2cm} \underline{$81.10\pm1.81$} \hspace{0.2cm} $75.20\pm2.32$ \newline  $55.90\pm0.98$ \hspace{0.2cm} $50.40\pm1.68$\hspace{0.2cm} \underline{$58.20\pm2.21$} \hspace{0.2cm} $54.33\pm2.45$\newline $48.60\pm1.96$ \hspace{0.2cm} $46.90\pm2.82$\hspace{0.2cm} \underline{$50.90\pm2.30$}\hspace{0.2cm} $48.41\pm2.57$\\
\hline
\end{tabular}

\end{table*}
\end{center}
\section{\textbf{Simulation Results}} \label{sec:EX}
In this section, we evaluate the performance of the proposed $\KFA$ scheme through a comprehensive set of experiments analyzing the model's performance under various settings, i.e., different data sizes and heterogeneity levels. 

\subsection{\textbf{Simulation Setup}}
\vspace{.025in}
\noindent
\textbf{\textit{Datasets:}} We conduct simulations using two image datasets: MNIST and CIFAR10. We use a Dirichlet distribution to model expected probabilities over a set of categories to account for the varying distribution of local data among clients. Dirichlet distribution is represented as $Dir(\alpha)$, where $\alpha$ adjusts the non-IID-ness degree, i.e., heterogeneity level.
As shown in Fig.~\ref{fig_data_het}, smaller values of $\alpha$ result in more skewed and, therefore,  more non-IID data.

\vspace{.025in}
\noindent
\textbf{\textit{Model Architecture:}} In our simulations, two distinct CNN architectures are utilized for MNIST and CIFAR-10 datasets. Specifically, we employ $M1 = [CU_1(32); CU_2(64); FC_1(64); FC_2(32); F_3(10)]$ for the MNIST dataset and  $M2 = [CU_1(16); CU_2(16); CU_3(32); CU_4(32); FC_1(128); FC_2(10)]$ for CIFAR-10, where $CU_m(t)$ represents the $m^{\text{th}}$ convolutional layer with $t$ channels, and $FC_m(t)$ signifies the $m^{\text{th}}$ dense layer with a size of $t$ neurons.

\vspace{.025in}
\noindent
\textbf{\textit{Baselines:}} The proposed $\KFA$ algorithm is compared with FedMD~\cite{FedMD}, Selective-FD~\cite{Selective knowledge Sharing}, and local training of the localized models on local datasets (referred to as Local). To ensure fairness across different methods, we maintain the size of the transfer set equal to the local data size. In knowledge fusion-based methods, i.e., $\KFA$, Selective-FD, and FedMD algorithms, local models are initially updated on their respective local datasets and then fine-tuned on the transfer set using ensemble knowledge. Conversely, in the Local method, both updating and fine-tuning phases occur solely on the local dataset without any knowledge sharing from other clients. 

Average Local Model Accuracy (ALMA) serves as the benchmark metric for all methods, reflecting the average test accuracy of all local models on their respective local test datasets.
The reported results represent the mean and standard deviation derived from three separate repetitions with distinct random seeds for local model initialization and distinct local data distributions. In each run, the local model initialization and local datasets are the same for all methods. 

\vspace{.025in}
\noindent
\textbf{\textit{Hyperparameters:}}
The local epoch is set to $E=1$ with a batch size of $128$, $64$, $32$, $16$, and $8$ samples for local data sizes of $1000$, $500$, $200$, $100$, and $50$, respectively. We consider $20$ clients in the FL system. The parameter $\beta$ in Eq.~\eqref{beta} is set to $10$ in all experiments. We execute several experiments for different heterogeneity levels, $\alpha = \{0.1, 0.25, 0.5, 1\}$, and various local data sizes, $|\mathbb D_n| = \{50, 100, 200, 500, 1000\}$.  

\subsection{\textbf{Simulation Results and Performance Analysis}}
In this section, our primary aim is to evaluate the effectiveness of various methods, specifically knowledge-sharing-based algorithms and the local training method, from the perspectives of local data size and heterogeneity level. Fig.~\ref{local data size} demonstrates the impact of different local data sizes on the performance, i.e., ALMA, of various algorithms for a fixed level of heterogeneity. Across all three scenarios, the proposed $\KFA$ algorithm exhibits superior ALMA compared to other baselines. Notably, when the local data size is set to $50$, the standard deviation of ALMA for the local training (Local) method varies considerably among different repetitions, although its average ALMA surpasses that of the FedMD algorithm. As the local data size increases, the performance gap between different methods narrows.

Fig.~\ref{heterogeneity level} depicts the performance of various methods across different levels of heterogeneity for a fixed local data size. In scenarios of high heterogeneity, i.e., strong non-IID scenarios with $\alpha = 0.1$, knowledge-sharing-based methods do not outperform the local training method. The $\KFA$ algorithm has superior or comparable ALMA compared to the baselines. However, as the non-IID degree of local datasets decreases, the performance of knowledge fusion-based methods converges and surpasses that of the local training method. In summary, in settings with large and highly heterogeneous local data, knowledge fusion algorithms do not offer advantages over the local training method.

Table~\ref{MNIST_table}  displays the ALMA (average $\pm$ standard deviation) metric of various methods on the MNIST datasets across different settings, encompassing various local data sizes and heterogeneity levels. Likewise, Table~\ref{CIFAR_table} presents the performance of baseline methods on the CIFAR-10 datasets under diverse settings. Unlike the MNIST dataset, the local training method demonstrates superior performance compared to knowledge-sharing-based methods in most scenarios, except for the setting where $\alpha = 1$ and $|\mathbb D_n| = 500$. This disparity may stem from the local models' inadequate performance, resulting in the generation of low-quality local knowledge. Similar to the results of the MNIST dataset, it can be observed that in conditions of large and highly heterogeneous local datasets, the local training method is preferable to knowledge fusion-based algorithms. However, the $\KFA$ algorithm outperforms the other knowledge fusion-based methods in most settings.

\section{\textbf{Conclusion}} \label{sec:CONC}
In conclusion, FL represents a significant shift from centralized ML approaches, providing a privacy-preserving, decentralized framework for training models across various nodes. Despite its advantages, FL faces challenges such as the requirement for uniform model architectures, and model drift due to non-IID local datasets. While FKD emerged as a solution, leveraging the KD concept to mitigate some of these issues, it fail to fully address model drift, highlighting the need for selective knowledge fusion. The proposed $\KFA$ algorithm offers a novel approach by evaluating and fusing only the relevant knowledge among clients, showcasing superior performance on MNIST and CIFAR10 datasets. This research underlines the potential of personalized knowledge fusion in managing the complexities of FL environments, particularly in the presence of diverse and heterogeneous data.

\end{document}